\documentclass[twoside]{article}
\usepackage{proceed2e}

\usepackage[latin2]{inputenc}
\usepackage{amsmath}
\usepackage{amssymb}
\usepackage{bm}
\usepackage[
    pdfstartview={FitH},
    bookmarks=true,
    bookmarksnumbered=true,
    bookmarksopen=true,
    bookmarksopenlevel=\maxdimen,
    pdfborder={0 0 0},
    colorlinks=true,
    linkcolor = black,
    citecolor=black,
    urlcolor=black,
    filecolor=black,
    pdfauthor={Ferenc Huszar and Simon Lacoste-Julien},
    pdftitle={A Kernel Approach to Tractable Bayesian Nonparametrics},
    pdfdisplaydoctitle=true
]{hyperref}
\usepackage[pdftex]{graphicx}
\usepackage[numbers,square]{natbib}
\usepackage{tikz}
\usetikzlibrary{arrows,shapes}

\newcommand{\ourmethod}{kst}
\newcommand{\figref}[1]{Fig.~\ref{#1}}
\newcommand{\subfigref}[2]{Fig.~\ref{#1}.#2}
\newcommand{\studentt}{student-$\mathcal{T}$\ }

\newcommand{\transpose}[1]{#1^{\mbox{\sf \scriptsize T}}}
\newcommand{\quadform}[2]{\transpose{#1}#2#1}

\newcommand{\varconc}{\alpha}
\newcommand{\meanconc}{\beta}
\newcommand{\ie}{i.\,e.\ }
\newcommand{\eg}{e.\,g.\ }

\title{A Kernel Approach to Tractable Bayesian Nonparametrics}

\author{Ferenc Husz\'{a}r\\University of Cambridge \And Simon Lacoste-Julien\\University of Cambridge}

\begin{document}

\maketitle

\begin{abstract}
Inference in popular nonparametric Bayesian models typically relies on sampling or other approximations. This paper presents a general methodology for constructing novel tractable nonparametric Bayesian methods by applying the kernel trick to inference in a parametric Bayesian model. For example, Gaussian process regression can be derived this way from Bayesian linear regression. Despite the success of the Gaussian process framework, the kernel trick is rarely explicitly considered in the Bayesian literature. In this paper, we aim to fill this gap and demonstrate the potential of applying the kernel trick to tractable Bayesian parametric models in a wider context than just regression. As an example, we present an intuitive Bayesian kernel machine for density estimation that is obtained by applying the kernel trick to a Gaussian generative model in feature space.
\end{abstract}

\section{Introduction}

The popularity of nonparametric Bayesian methods has steadily risen in machine learning over the past decade. Bayesian inference in almost all current nonparametric models relies on approximations which typically involve Markov chain Monte Carlo, or more recently variational approximations~\cite{Blei2004}. There is an ever-growing interest in developing nonparametric Bayesian methods in which inference and prediction can be expressed in closed form and no approximations are needed. Such methods would be quite useful and desired in many applications, but are unfortunately rare at present. Perhaps the only known example is Gaussian process (GP) regression. In GP regression~\cite{Rasmussen2006}, the posterior predictive distribution is a tractable Gaussian with parameters that can be computed from data exactly in polynomial time. The method is widely adopted and also has favourable frequentist asymptotic properties~\cite{vanderVaart2008}. But what is the secret behind the remarkable algorithmic clarity of GP regression? What makes closed-form computations possible? We argue that the key is that GP regression is also a \emph{kernel machine}: the method is arrived at by applying the kernel trick in a parametric Bayesian model, namely linear regression (see \eg Chapter~2 in~\cite{Rasmussen2006})

Kernel methods~\cite{Hofman2008,Scholkopf2002} use a simple trick, widely known as the kernel trick, to overcome the limitations of a linear model: the observations ${\bm{x}_i\in\mathcal{X}}$ are first embedded in a feature space $\mathcal{F}$ using a nonlinear mapping ${\varphi:\mathcal{X} \mapsto \mathcal{F}}$. A linear algorithm is then applied on the embedded representations $\bm{\phi}_n = \varphi(\bm{x}_n)$ instead of the observations themselves. If the algorithm only makes use of scalar products $\langle\bm{\phi}_n,\bm{\phi}_m\rangle$, then by replacing all scalar products by tractable \emph{kernel} evaluations ${k(\bm{x}_n,\bm{x}_m) := \langle\bm{\phi}_n,\bm{\phi}_m\rangle}$, the expressive power can be substantially increased with only a minor increase in computational costs. Notably, the kernel trick allows one to construct nonparametric machine learning methods from parametric ones.

Despite its popularity in ``non-Bayesian'' studies, the kernel trick is rarely considered as a construction tool in Bayesian nonparametrics. GP regression is a rare, if not the only example. GPs are therefore often called \emph{Bayesian kernel machines}. In this paper,  we consider finding new examples of Bayesian kernel machines, \ie broadening the intersection between kernel machines and nonparametric Bayesian methods. For Bayesian nonparametrics, the kernel approach offers invaluable closed-form computations and a rigorous analysis framework associated with reproducing kernel Hilbert spaces (RKHS). For kernel machines, a Bayesian formulation offers benefits, such as novel probabilistic approaches to setting kernel hyperparameters and means of incorporating such methods in hierarchical Bayesian models~\cite{Rasmussen2002,Chu2007,Adams2009}.

In this paper, we present a methodology for finding novel Bayesian kernel machines, following the recipe of GP regression:
\begin{enumerate}
\addtolength{\itemsep}{-2mm}
 \item Start with a simple Bayesian model of observations.
 \item Derive exact Bayesian inference in the model.
 \item Express the posterior predictive distribution in terms of dot products and apply the kernel trick.
\end{enumerate}

The crucial point is finding the basic model in step 1, in which both steps 2 and 3 are possible. Fortunately, this search is guided by intuitive orthonormal invariance considerations that will be demonstrated in this paper. We present an example Bayesian kernel machine for density estimation, based on the linear Gaussian generative model underlying principal component analysis (PCA)~\cite{Roweis1999}. We show that the kernel trick can be applied in the Bayesian method by choosing prior distributions over the parameters which preserve invariance to orthonormal transformations.

The rest of the paper is organised as follows. In Sec.~\ref{sec:genmodel}, we review Bayesian inference in a Gaussian generative model and discuss consequences of applying the kernel trick to the predictive density. We consider the infinite dimensional feature spaces case in Subsec.~\ref{sec:infinite_dimensional}. In Sec.~\ref{sec:experiments}, we present experiments on high dimensional density estimation problems comparing our method to other Bayesian and non-Bayesian nonparametric methods.

\section{A Gaussian model in feature space\label{sec:genmodel}}

Assume that we have observations $\bm{x}_i$ in a $d$-dimensional Euclidean space and that our task is to estimate their density. In anticipation of the sequel, we embed the observations $\bm{x}_i$ into a high-dimensional feature space $\mathcal{F}$ with an injective smooth nonlinear mapping ${\varphi:\mathcal{X} \mapsto \mathcal{F}}$. Our density estimation method is based on a simple generative model on the ambient feature space of the embedded observations $\bm{\phi}_i=\varphi(\bm{x}_i)$. For now, think of $\mathcal{F}$ as a $D$-dimensional Euclidean space and $\bm{\phi}_i$ as \emph{arbitrary elements} of $\mathcal{F}$ (\ie they are not necessary constrained to lie on the \emph{observation manifold} $\mathcal{O} = \left\{\varphi(\bm{x}) : \bm{x}\in\mathcal{X}\right\}$). We suppose that $\bm{\phi}_i$ were sampled from a Gaussian distribution with unknown mean $\bm{\mu}$ and covariance $\bm{\Sigma}$:
\begin{equation}
 \bm{\phi}_{1:N}\vert \bm{\mu},\bm{\Sigma} \sim \mathcal{N}\left(\bm{\mu},\bm{\Sigma}\right)\mbox{, i.\,i.\,d.}\label{eqn:genmodel_obs}
\end{equation}
The notation $\bm{\phi}_{1:N}$ is used to denote the set of vectors $\bm{\phi}_{1}$ up to $\bm{\phi}_{N}$.


Now we will consider estimating parameters of this model in a Bayesian way. In Bayesian inference, one defines prior distributions over the parameters and then uses Bayes' rule to compute the posterior over them. Importantly, now we also require that the resulting Bayesian procedure is still amenable to the kernel trick. We therefore start by discussing a necessary condition for kernelisation which can then guide our choice of prior distributions.

The kernel trick requires that the algorithm be expressed solely in terms of scalar products $\langle\bm{\phi}_i,\bm{\phi}_j\rangle$. Scalar products are invariant under orthonormal transformations of the space, \ie if $\mathcal{A}$ is an orthonormal transformation then $\left\langle \bm{u},\bm{v} \right\rangle = \left\langle \mathcal{A}\bm{u},\mathcal{A}\bm{v} \right\rangle$ for all $\bm{u}$, and $\bm{v}$ in the space. Thus, if one wants to express an algorithm in terms of scalar products, it has to be -- at least -- invariant under orthonormal transformations, such as
rotations, reflections and permutations. It is well known that PCA has this property, but, for example, factor analysis (FA) does not~\cite{Roweis1999}, thus one cannot expect a kernel version of FA without any restrictions.

Another desired property of the method is analytical convenience and tractability, which can be ensured by using conjugate priors. The conjugate prior of the Gaussian likelihood is the \emph{Normal-inverse-Wishart}, which in our case has to be restricted to meet the orthonormal invariance condition:
\begin{equation} \label{eqn:genmodel_mean}
\begin{split}
 \bm{\Sigma};\:\sigma_0^{2}, \varconc &\sim \mathcal{W}^{-1}\left(\sigma^{2}_0\bm{I},\varconc\right)  \\ \bm{\mu}\vert\bm{\Sigma},\meanconc &\sim \mathcal{N}\left(\bm{0},\frac{1}{\meanconc}\bm{\Sigma}\right) ,
\end{split}
\end{equation}
where $\mathcal{W}^{-1}$ and $\mathcal{N}$ denote the inverse-Wishart and Gaussian distributions with the usual parametrisation. The general Normal-inverse-Wishart family had to be restricted in two ways: firstly, the mean of $\bm{\mu}$ was set to zero; secondly, the scale matrix of the inverse-Wishart was set to be spherical. These sensible restrictions ensure that the marginal distribution of $\bm{\phi}$ is centered at the origin and is spherically symmetric and therefore orthonormal invariance holds, which is required for kernelisation.

Having defined the hierarchical generative model in eqns.\ \eqref{eqn:genmodel_obs}--\eqref{eqn:genmodel_mean}, our task is to estimate the density of $\bm{\phi}$'s given the previous observations $\bm{\phi}_{1:N}$, which in a Bayesian framework is done by calculating the following posterior predictive distribution:
\begin{multline*}
 p(\bm{\phi}\vert\bm{\phi}_{1:N};\sigma_0^{2},\varconc,\meanconc) = \\
 \int p(\bm{\phi}\vert \bm{\mu},\bm{\Sigma})p(\bm{\mu},\bm{\Sigma}\vert\bm{\phi}_{1:N};\sigma_0^{2},\varconc,\meanconc) d\bm{\mu}d\bm{\Sigma}
\end{multline*}

By straightforward computation, it can be shown that the posterior predictive distribution is a $D$-dimensional \studentt distribution of the form (with the dependence on hyper-parameters made implicit):
\begin{align}
  &p(\bm{\phi}\vert\bm{\phi}_{1:N}) \propto \label{eqn:postpredictive} \\
  &\left( \gamma + \quadform{\tilde{\bm{\phi}}}{\left(\sigma_0^{2}\bm{I} + \tilde{\bm{\Phi}}\left(\bm{I} - \frac{\bm{1}\transpose{\bm{1}}}{N+\meanconc}\right)\transpose{\tilde{\bm{\Phi}}}\right)^{-1}}\right)^{-\frac{1 + N + \varconc}{2}} . \notag
\end{align}
where $\gamma = \frac{1+\meanconc+N}{\meanconc+N}$,  $\tilde{\bm{\phi}} = \bm{\phi} - \frac{N\bar{\bm{\phi}}}{N+\meanconc}$, $\tilde{\bm{\Phi}} = \left[\bm{\phi}_1 - \bar{\bm{\phi}},\ldots,\bm{\phi}_N - \bar{\bm{\phi}}\right]$, $\bar{\bm{\phi}} = \frac{1}{N}\sum_{n=1}^{N}\bm{\phi}_n$ is the empirical mean in feature space and $\bm{1}$ is a $N\times1$ vector of ones.

In order to obtain an expression which only contains scalar products, we invoke Woodbury's matrix inversion formula~\cite{Hager1989}:
\begin{align}
  &p(\bm{\phi}\vert\bm{\phi}_{1:N}) \propto \left( \gamma + \frac{\quadform{\tilde{\bm{\phi}}}{}}{\sigma_0^2} \:- \right. \label{eqn:postpredictive_Woodbury} \\
  &\left. \quadform{\tilde{\bm{\phi}}}{ \tilde{\bm{\Phi}} \left(\sigma_0^{4}\left(\bm{I} + \frac{\bm{1}\transpose{\bm{1}}}{\meanconc}\right) + \sigma_0^2\quadform{\tilde{\bm{\Phi}}}{}\right)^{-1}\transpose{\tilde{\bm{\Phi}}}} \right)^{-\frac{1 + N + \varconc}{2}}. \notag
\end{align}

\subsection{Kernel trick}

Until now, we assumed that $\bm{\phi}$ was an arbitrary point in $D$-dimensional space. In reality, however, we only want to assign probabilities~\eqref{eqn:postpredictive_Woodbury} to points on the so-called observation manifold, $\mathcal{O} = \left\{\varphi(\bm{x}) : \bm{x}\in\mathcal{X}\right\}$, \ie to points that can be realised by mapping an observation $\bm{x}\in\mathcal{X}$ to feature space $\bm{\phi} = \varphi(\bm{x})$. Restricted to $\mathcal{O}$, we can actually make use of the kernel trick, assuming as well that $\bm{\phi}_i=\varphi(\bm{x}_i)$ for the previous observations. Indeed, Eq.~\eqref{eqn:postpredictive_Woodbury} then only depends on $\bm{x}_{1:N}$ and $\bm{x}$ through $\quadform{\tilde{\bm{\phi}}}{}$, $\transpose{\tilde{\bm{\phi}}}\tilde{\bm{\Phi}}$ and $\quadform{\tilde{\bm{\Phi}}}{}$, which can all be expressed in terms of pairwise scalar products $\transpose{\bm{\phi}_n}\bm{\phi}_m = \langle\varphi(\bm{x}_n),\varphi(\bm{x}_m)\rangle = k(\bm{x}_n,\bm{x}_m)$ as follows:

\begin{align}
 \quadform{\tilde{\bm{\phi}}}{} &= k(\bm{x},\bm{x}) - 2\frac{\sum_i k(\bm{x},\bm{x}_i)}{N + \meanconc} + \frac{\sum_{i,j} k(\bm{x}_i,\bm{x}_j)}{\left(N + \meanconc\right)^{2}} \label{eqn:kernelexpressions} \\
 \left[\transpose{\tilde{\bm{\phi}}}\tilde{\bm{\Phi}}\right]_{n} &=
     \begin{aligned}[t]
     k(\bm{x},\bm{x}_n) &- \frac{\sum_i k(\bm{x},\bm{x}_i)}{N} \\
     &+ \frac{\sum_{i,j} k(\bm{x}_i,\bm{x}_j) - N\sum_i k(\bm{x}_n,\bm{x}_i)}{N\left(N+\meanconc\right)} \\
     \end{aligned} \notag \\
 \left[\quadform{\tilde{\bm{\Phi}}}{}\right]_{n,m} &=
    \begin{aligned}[t]
    k(\bm{x}_n,\bm{x}_m) &- \frac{\sum_i k(\bm{x_n},\bm{x}_i) + \sum_i k(\bm{x_m},\bm{x}_i)}{N} \\
    &+ \frac{\sum_{i,j} k(\bm{x}_i,\bm{x}_j)}{N^2}
    \end{aligned} \notag
\end{align}

As said previously, we are only interested in points in the (curved) observation manifold. The restriction of the predictive distribution~\eqref{eqn:postpredictive_Woodbury} to the observation manifold induces the same density function $q(\bm{\varphi}(\bm{x}) \vert \bm{x}_{1:N}) = p(\bm{\varphi}(\bm{x}) \vert\bm{\phi}_{1:N})$ (which is now unnormalised), but with respect to a $d$-dimensional Lebesgue base measure in the geodesic coordinate system of the manifold. Finally, we map the density $q(\bm{\varphi}(\bm{x}) \vert \bm{x}_{1:N})$ back to the original input space by implicitly inverting the mapping $\varphi$, yielding the (unnormalised) predictive density $q_k(\bm{x} \vert \bm{x}_{1:N})$ defined on the input space. By doing so, we need to include a multiplicative Jacobian correction term which relates volume elements in the tangent space of the manifold to volume elements in the input space $\mathcal{X}$:
\begin{multline} \label{corrected_density}
q_k(\bm{x} \vert \bm{x}_{1:N}) = \\
q(\bm{\varphi}(\bm{x})\vert\bm{x}_{1:N}) \cdot \det\left( \left\langle\frac{\partial\varphi(\bm{x})}{\partial x^i} ,  \frac{\partial\varphi(\bm{x})}{\partial x^j}\right\rangle \right)^{\frac{1}{2}}_{i,j=1,\ldots, d} .
\end{multline}
This formula can be be derived from the standard change-of-variable formula for densities, as we show in the Appendix \ref{sec:change_of_variable}.

\begin{figure*}
    \begin{center}
        \begin{tikzpicture}
            \node[anchor = south west, inner sep = 0in,] at (10pt,10pt) {\includegraphics[trim = 1.3in 2.4in 0.93in 2in, clip, width=6.45in,keepaspectratio]{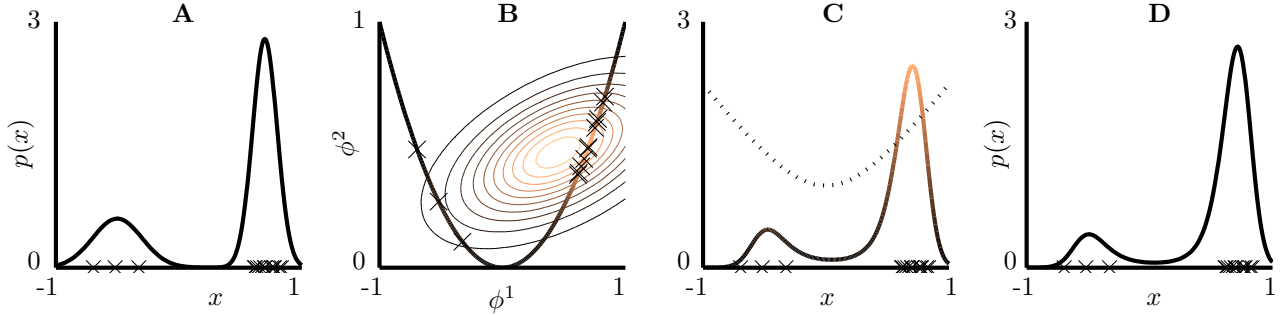}};
            \def\offsetB{1.70in}
            \def\offsetC{3.40in}
            \def\offsetD{5.11in}

            \node at (0.85in,1.5in) {\textbf{A}};

            \node at (10pt,5pt) {-1};
            \node at (1.02in,0in) {$x$};
            \node at (1.43in,5pt) {1};

            \node at (5pt,15pt) {0};
            \node at (0in,0.82in) [rotate = 90]{$p(x)$};
            \node at (5pt,1.47in) {3};

            \node at (\offsetB+0.85in,1.5in) {\textbf{B}};

            \node at (\offsetB+10pt,5pt) {-1};
            \node at (\offsetB+0.82in,0in) {$\phi^1$};
            \node at (\offsetB+1.43in,5pt) {1};

            \node at (\offsetB+5pt,15pt) {0};
            \node at (\offsetB+0in,0.82in) [rotate = 90]{$\phi^2$};
            \node at (\offsetB+5pt,1.47in) {1};

            \node at (\offsetC+0.85in,1.5in) {\textbf{C}};

            \node at (\offsetC+10pt,5pt) {-1};
            \node at (\offsetC+0.82in,0in) {$x$};
            \node at (\offsetC+1.47in,5pt) {1};

            \node at (\offsetC+5pt,15pt) {0};
            \node at (\offsetC+5pt,1.47in) {3};

            \node at (\offsetD+0.85in,1.5in) {\textbf{D}};

            \node at (\offsetD+10pt,5pt) {-1};
            \node at (\offsetD+0.82in,0in) {$x$};
            \node at (\offsetD+1.43in,5pt) {1};

            \node at (\offsetD+5pt,15pt) {0};
            \node at (\offsetD+0in,0.82in) [rotate = 90]{$p(x)$};
            \node at (\offsetD+5pt,1.47in) {3};
        \end{tikzpicture}
    \end{center}

    \caption{
        \label{fig:illustration_2D}
        Illustration of kernel \studentt density estimation on a toy problem.
        \textbf{A,\ }Fourteen data points $x_{1:14}$ (\emph{crosses}) drawn from from a mixture distribution (\emph{solid curve}).
        \textbf{B,\ }Embedded observations $\bm{\phi}_{1:14}$(\emph{crosses}) in the two dimensional feature space and contours of the predictive density $q(\bm{\phi}_{15}\vert x_{1:14})$. The observation manifold is a parabola (\emph{solid curve}), and we are interested in the magnitude of the predictive distribution to this manifold (\emph{colouring}).
        \textbf{C,\ }The restricted posterior predictive distribution pulled back to the real line (\emph{solid curve}) gets multiplied by the Jacobian term $\sqrt{1 + 4x^2}$ (\emph{dotted curve}) to give
        \textbf{D,\ }the predictive distribution $q_{k}(x_{15}\vert x_{1:14})$ (\emph{solid curve}) in observation space.
         All distributions are scaled so that they normalise to one.}
\end{figure*}

So what exactly did we gain by applying the kernel trick? Despite the fact that the simple density model in the whole feature space is unimodal, in the local coordinate system of the non-linear observation manifold, the predictive density may appear multimodal and hence possess interesting nonlinear features. This is illustrated in Fig.~\ref{fig:illustration_2D}. Assume that we observe draws from a complicated distribution, which cannot be conveniently modelled with a linear Gaussian model (Fig.~\ref{fig:illustration_2D}.A). We map observations to a two dimensional feature space using the mapping $\varphi^1(x)=x,\varphi^2(x)=(x)^2$ hoping that the embedded observations are better fitted by a Gaussian, and carry out inference there (Fig.~\ref{fig:illustration_2D}.B). Note how all possible observations in the original space get mapped to the observation manifold~$\mathcal{O}$, which is now the parabola $\phi^2=(\phi^1)^2$. The outcome of inference is a posterior predictive distribution in feature space, which takes a \studentt form. To obtain the predictive distribution in the observation space we have to look at the magnitude of the predictive distribution along the observation manifold. This is illustrated by the contour lines in panel B. The restricted predictive distribution is then ``pulled back'' to observation space and has to be multiplied by the Jacobian term (Fig.~\ref{fig:illustration_2D}.C) to yield the final estimation of the density of the observed samples (Fig.~\ref{fig:illustration_2D}.D). Remarkably, our method computes the unnormalised density estimate (before the Jacobian term is added, Fig.~\ref{fig:illustration_2D}.C) directly from the data (Fig.~\ref{fig:illustration_2D}.A). All intermediate steps, including the inversion of the mapping, are \emph{implicit} in the problem formulation, and we never have to work with the feature space directly. As the method essentially estimates the density by a \studentt in feature space, we shall call it \emph{kernel \studentt density estimation}.

\subsection{Infinite dimensional feature spaces \label{sec:infinite_dimensional}}

Of course, we constructed the previous toy example so that the simple normal model in this second order polynomial feature space is able to model the bimodal data distribution. Should the data distribution be more complicated, \eg would have three or more modes, the second order features would hardly be enough to pick up all relevant statistical properties. Intuitively, adding more features results in higher potential expressive power, while the Bayesian integration safeguards us from overfitting. In practice, therefore, we will use the method in conjunction with rich, perhaps infinite dimensional feature spaces, which will allow us to relax these parametric restrictions and to apply our method to model a wide class of probability distributions.

However, moving to infinite dimensional feature spaces introduces additional technical difficulties. As for one, Lebesgue measures do not exist in these spaces, therefore our derivations based on densities with respect to Euclidean base measure should be reconsidered. For a fully rigorous description of the method in infinite spaces, one may consider using Gaussian base measures instead. In this paper, we address two specific issues that arise when the feature space is large or infinite dimensional.

Firstly, because of the Jacobian correction term, the predictive density $q_k$ in input space is still not fully kernelised in~\eqref{corrected_density}. We note though that this term is the determinant of a small $d\times d$ matrix of inner products, which can actually be computed efficiently for some kernels. For example, the all-subsets kernel~\cite{Cristianini2004} has a feature space with exponential dimension $D=2^d$, but each inner product of derivatives can be computed in $\mathcal{O}(d)$. But what about nonparametric kernels with infinite $D$? Interestingly, we found that shift invariant kernels -- for which $k(\bm{x},\bm{y}) = k(\bm{x}+\bm{z},\bm{y}+\bm{z})$ for all $\bm{z}$ -- actually yield a \emph{constant Jacobian term} (see proof in the Appendix \ref{sec:Jacobian_shift}) so it can be safely ignored to obtain a fully kernelised -- although unnormalised -- predictive density $q_k(\bm{x} \vert \bm{x}_{1:N}) = q(\bm{\varphi}(\bm{x})\vert\bm{x}_{1:N})$. 
This result is important, as the most commonly used nonparametric kernels, such as the squared exponential, or the Laplacian fall into this category.

The second issue is normalisation: Eq.~\eqref{corrected_density} only expresses the predictive distribution up to a multiplicative constant. Furthermore, the derivations implicitly assumed that $\alpha>D-1$, where $D$ is the dimensionality of the feature space, otherwise the probability densities involved become improper. This assumption clearly cannot be satisfied when $D$ is infinite.
However, one can argue that even if the densities in the feature space are improper, the predictive distributions that we use may still be normalisable when restricted to the observation manifold for $\alpha>d-1$. Indeed, the two-dimensional \studentt is normalisable when restricted to the parabola as in Fig.~\ref{fig:illustration_2D} for all $\alpha>0$ rather than $\alpha>1$ (to show this, consider $N=0$ so that~\eqref{eqn:postpredictive} takes the simple form $q(\bm{\phi}) = (\textrm{cnst.}+ \transpose{\bm{\phi}}\bm{\phi})^{-\frac{1+\alpha}{2}}$). So although distributions in feature space may be improper, the predictive densities may remain well-defined even when nonparametric kernels are used.

Given these observations, we thus decide to \emph{formally} apply equation~\eqref{corrected_density} for our predictive density estimation algorithm, ignoring the normalisation constant and also the Jacobian term when shift-invariant nonparametric kernels are applied.

\begin{figure*}[t]
 \begin{center}
  \begin{tikzpicture}
    \node[anchor = south west, inner sep = 0in] at (0in,4pt) {\includegraphics[width=2.1in,keepaspectratio]{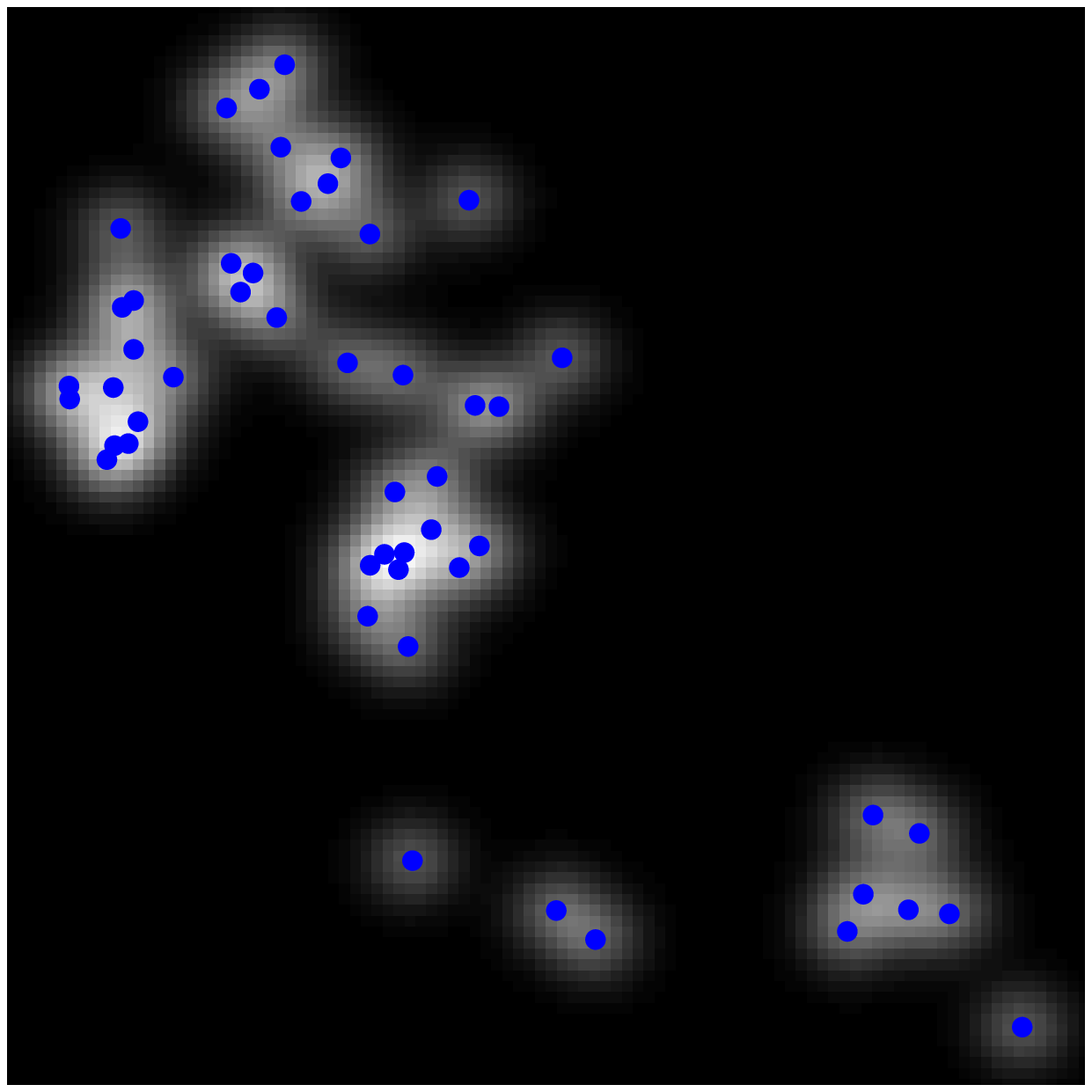}};
    \draw[ultra thick,draw = red] (0in,0pt) -- (0.15in,0pt);
    \node[inner sep = 0in] at (1.05in,2.25in) {\textbf{A}: $\alpha = 0.01,\beta = 0.01$};
    \node[anchor = south west, inner sep = 0in] at (2.15in,4pt) {\includegraphics[width=2.1in,keepaspectratio]{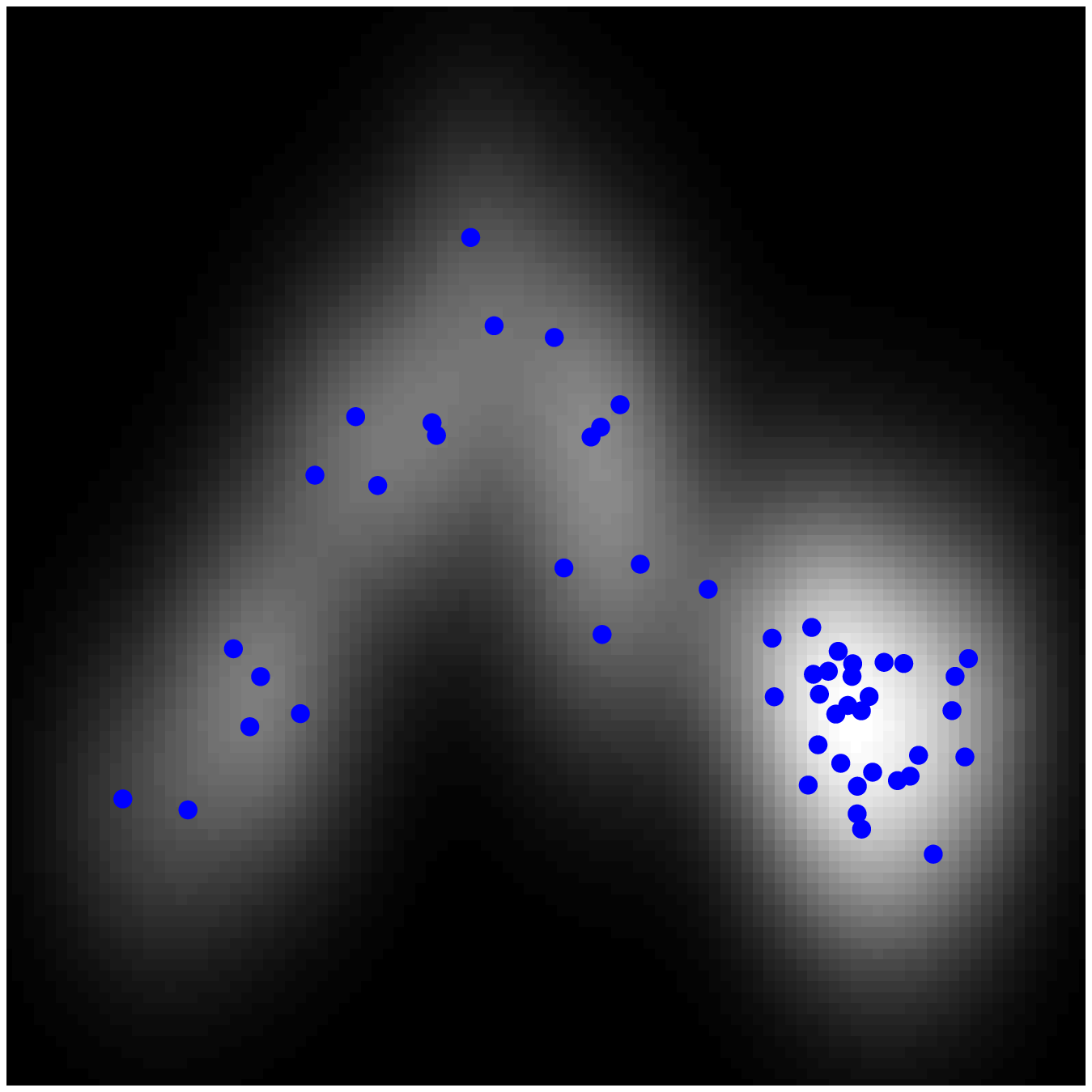}};
    \draw[ultra thick,draw = red] (2.15in,0pt) -- (2.85in,0pt);
    \node[inner sep = 0in] at (3.2in,2.25in) {\textbf{B}: $\alpha = 3,\beta = 0.01$};
    \node[anchor = south west, inner sep = 0in] at (4.3in,4pt) {\includegraphics[width=2.1in,keepaspectratio]{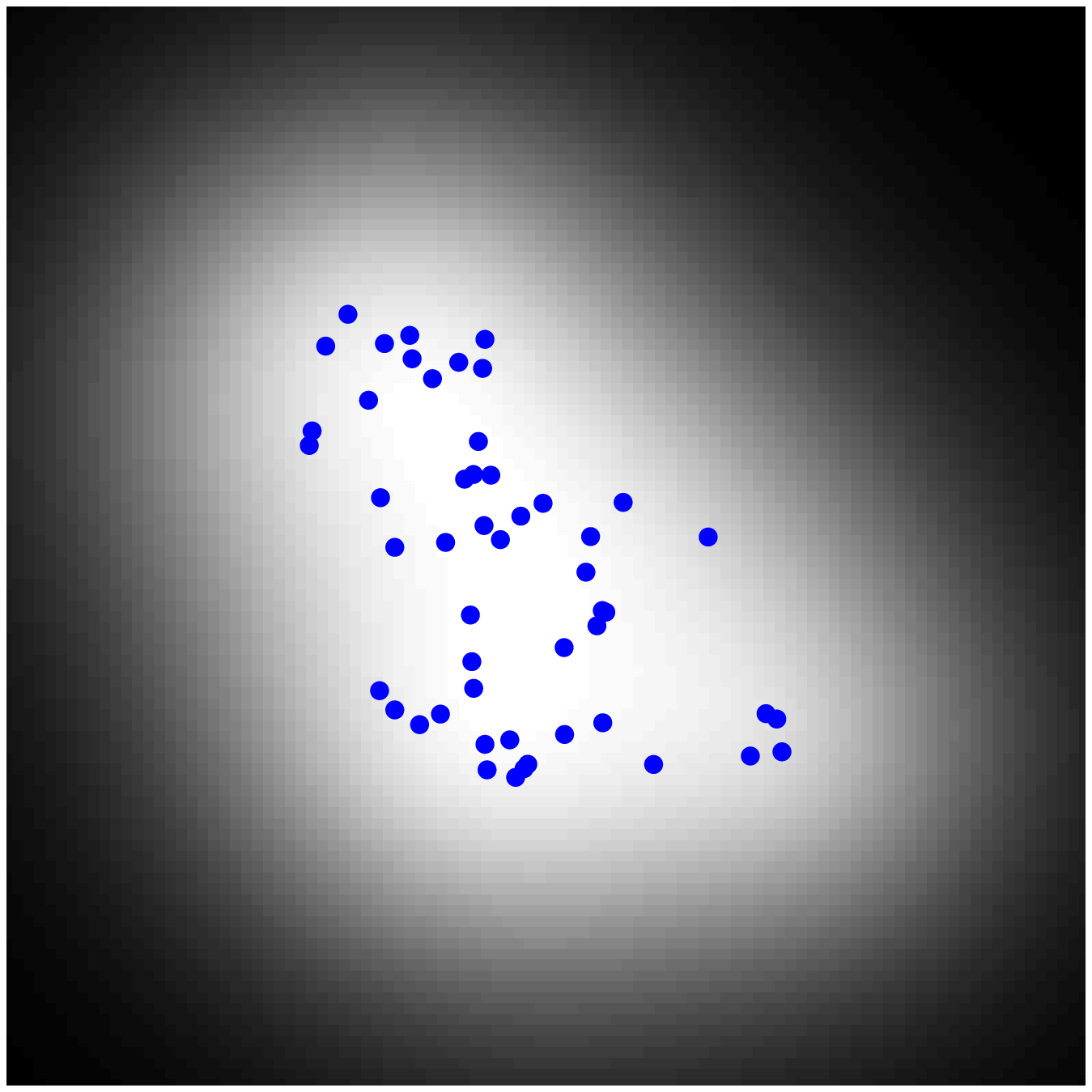}};
    \draw[ultra thick,draw = red] (4.3in,0pt) -- (6in,0pt);
    \node[inner sep = 0in] at (5.35in,2.25in) {\textbf{C}: $\alpha = 10, \beta = 0.01$};
  \end{tikzpicture}
 \end{center}
 \caption{\label{fig:fantasy}Fantasy data drawn from a two dimensional kernel \studentt model with SE kernel. The  length-scale $\ell=1$ (\emph{bars below panels}) and $\meanconc=0.01$ are fixed, $\varconc$ varies between $0.01$ and $10$ (\textbf{A} to \textbf{C}). Panels show 50 points (\emph{blue dots}) drawn from the generative model and the predictive density (\emph{gray level}) of the $51^{\mbox{st}}$ draw.}
\end{figure*}

Finally, we want to emphasize that the method does not imply nor does it require that arbitrary distributions, when embedded in rich Hilbert spaces, will look exactly like Gaussian or \studentt measures. Instead, the method uses the potentially infinitely flexible family of Gaussian measures to approximately model distributions in the feature space, and uses Bayesian integration to address model-misspecification via explicit representation of uncertainty. As demonstrated in Fig.~\ref{fig:illustration_2D}, the richness and flexibility of the density model stems from projecting the distribution onto a nonlinear observation manifold, allowing us to overcome limitations of Gaussian measures, such as unimodality.

\section{Related work}

Our method is closely related to kernel principal component analysis~\cite[kPCA]{Scholkopf1998} as both of them essentially use the same underlying generative model (Eq.~\ref{eqn:genmodel_obs}). While our method is based on Bayesian inference, kPCA performs constrained maximum likelihood estimation~\cite{Rosipal2001} of parameters $\bm{\mu}$ and $\bm{\Sigma}$. kPCA is a very effective and popular method for nonlinear feature extraction. Its pitfall from a density estimation perspective is that it does not generally induce a sensible probability distribution in observation space. To understand this, consider performing kPCA of order one (\ie recover just a single nonlinear component) on the toy data in~\subfigref{fig:illustration_2D}{A}. The solution defines a degenerate Gaussian distribution in feature space which is concentrated~ along a line, roughly the long axes of the equiprobability ellipses in \subfigref{fig:illustration_2D}{B}. As such a distribution can intersect the observation manifold at most twice, the predictive distribution in observation space will be a mixture of two delta distributions. This degeneracy can be ruled out by recovering at least as many nonlinear components as the dimensionality of feature space, but this is clearly not a viable option in infinite or large dimensional feature spaces. The Bayesian approach sidesteps the degeneracy problem by integrating out the mean and covariance, thereby averaging many, possibly degenerate distributions to obtain a non-degenerate, smooth \studentt distribution in feature space, which results in a smooth density estimate in observation space.

Another particularly interesting related topic is that of \emph{characteristic kernels}~\cite{Sriperumbudur2008}. A positive definite kernel $k(x,x') = \langle\varphi(x),\varphi(x')\rangle$ is called characteristic if the mean element $\mu_\pi = \mathbb{E}_{\mathbb{X} \sim \pi}[\varphi(\mathbb{X})]$ uniquely characterises any probability measure $\pi$. From a density estimation perspective, this suggests that estimating a distribution in observation space boils down to estimating the mean in a characteristic kernel space. The kernel moment matching framework~\cite{Song2008} tries to exploit characteristic kernels in a very direct way: parameters of an approximating distribution are chosen by minimising maximum mean discrepancy in feature space. Our method can be interpreted as performing Bayesian estimation of first and second moments in feature space, therefore one may hope that work on characteristic kernels will help us further study and understand properties of the algorithm.

The Gaussian process has been used as a nonparametric prior over functions for unsupervised learning tasks in several studies, such as in Gaussian process latent variable models~\cite[GPLVMs]{Lawrence2004} or the Gaussian process density sampler~\cite[GPDS]{Adams2009}. What sets the present approach apart from these is that while they incorporated Gaussian processes as a building block in an unsupervised model, we re-used the construction of GP regression and explicitly applied the kernel trick in an parametric unsupervised method. The GPDS method relies on Markov chain Monte Carlo (MCMC) methods for so called doubly-intractable distributions~\cite{Murray2006}. These methods can be used to sample from the posterior over hyper-parameters of probabilistic methods where normalisation is intractable. In particular, they could be used to integrate out the hyper-parameters $\varconc,\meanconc$ and $\sigma_0$, of our method. We emphasise that while MCMC is a crucial component of the GPDS, inference in our model is essentially closed-form.

\section{Experiments\label{sec:experiments}}

A common way to examine properties of unsupervised learning models is to draw \emph{fantasy datasets} from the generative model. The kernel \studentt model can be simulated as follows: first, we fix $\bm{x}_1=\bm{0}$ for convenience. Our generative model with shift-invariant kernels define an improper prior over the whole space when there is no observation (as it should since only difference between points is meaningful for shift invariant kernels). Then we draw each subsequent $\bm{x}_n$ from the kernel \studentt predictive distribution $q_{k}(\bm{x}_n\vert\bm{x}_{1:n-1};\sigma_o,\varconc,\meanconc)$ conditioned on previous draws. Sampling from the predictive distribution is possible via hybrid Monte Carlo~\cite{Neal2010}.

Fig.~\ref{fig:fantasy} shows example fantasy datasets for different settings of parameters. We observe that the generative process has an approximate clustering property, in which it behaves similarly to the Chinese restaurant process. We also note that the ``checkerboard - like'' eigenfunctions of the squared exponential kernel induce a tree like refinement of the space, in which the method is similar to Dirichlet diffusion trees.

\begin{figure}[t]
	\begin{center}
	\resizebox{\columnwidth}{!}{
	\begin{tikzpicture}
		\node[anchor = south west, inner sep = 0in] at (18pt,16pt) {\includegraphics[width=0.88\columnwidth,trim = 1.2in 0.8in 0.9in 0.6in, clip]{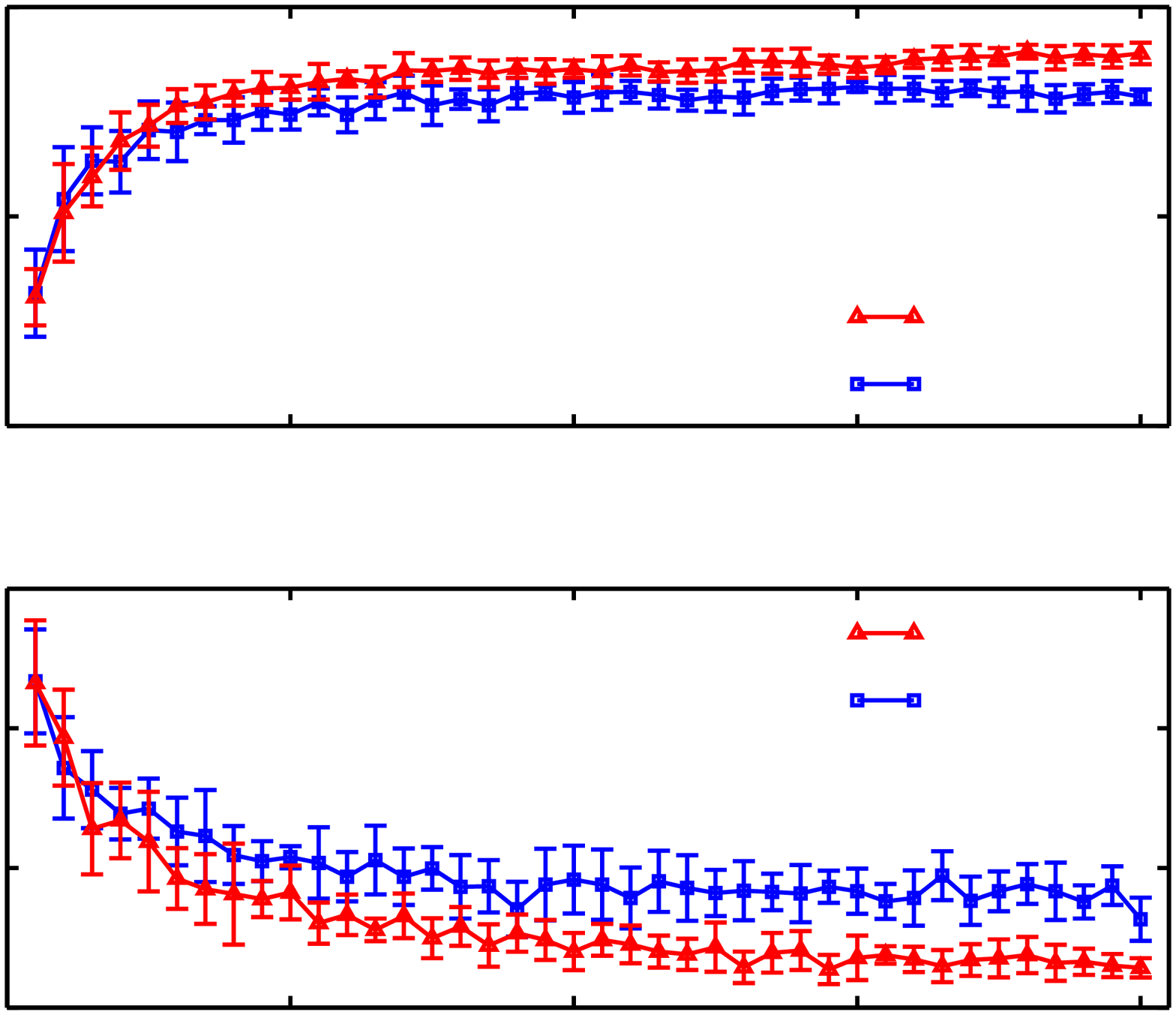}};
		\node[rotate=90] at (-3pt,0.76in) {confusion$[\%]$};
		\node[rotate=90] at (-3pt,2.15in) {AuC$[\%]$};
		\node at (1.63in,0pt) {number of training data};
		\node[anchor = east] at (20pt,0.28in) {0};
		\node[anchor = east] at (20pt,0.60in) {25};
		\node[anchor = east] at (20pt,0.92in) {50};
		\node[anchor = east] at (20pt,1.24in) {75};
		\node[anchor = west] at (2.5in,1.18in) {\emph{\ourmethod}};
		\node[anchor = west] at (2.5in,1.01in) {\emph{kde}};
		\node at (1.63in,1.4in) {label reconstruction};

		\node[anchor = east] at (20pt,1.65in) {50};
		\node[anchor = east] at (20pt,2.15in) {75};
		\node[anchor = east] at (20pt,2.65in) {100};
		\node[anchor = west] at (2.5in,1.94in) {\emph{\ourmethod}};
		\node[anchor = west] at (2.5in,1.77in) {\emph{kde}};
		\node at (1.63in,2.8in) {novelty detection};
	
		\node at (20pt,10pt) {0};
		\node at (0.94in,10pt) {100};
		\node at (1.63in,10pt) {200};
		\node at (2.32in,10pt) {300};
		\node at (3.01in,10pt) {400};
	\end{tikzpicture}
	}
	\end{center}
	\caption{\label{fig:AUC_and_Confusion} Performance of \emph{\ourmethod} and \emph{kde} as a function of the number of training examples  on USPS data. Errorbars show $\pm$ one standard deviation.}
\end{figure}

\begin{table}[t]
 \begin{center}
 \resizebox{\columnwidth}{!}{
  \begin{tabular}{|l||c|c|}
	\hline \textsc{method}& \textsc{novelty}(AuC[\%]) & \textsc{recons.}(conf[\%])\\
	\hline\hline \emph{\ourmethod}	& $\bm{96.82}\pm0.8$ & $\bm{2.15}\pm{1.0}$\\
	\hline	\emph{kde}	& $94.56\pm0.6$ & $9.24\pm1.9$\\
	\hline	\emph{dpm}	& $73.56\pm4.3$ & $23.85\pm8.2$\\
	\hline
  \end{tabular}
 }
 \end{center}
 \caption{\label{tab:digits_results}Performance of \emph{\ourmethod}, \emph{kde} and \emph{dpm} in novelty detection (\emph{novelty, left}) and label reconstruction (\emph{recons., right}) on the USPS data. The table shows average AuC and confusion values $\pm$ one standard deviation measured over 10 randomised experiments.}
\end{table}

The performance of density models where the normalisation constants are not available -- similar examples in the past included Markov random fields~\cite{Haluk1987} and deep belief nets -- is hard to evaluate directly. These density models are typically used in unsupervised tasks, such as reconstruction of missing data, novelty detection and image denoising, where the important quantity is the shape of the distribution, rather than the absolute density values.  In the following, we consider three such unsupervised tasks that are indicative of the model's performance: novelty detection, assessment of reconstruction performance and the recently introduced task of relative novelty detection~\cite{Smola2009}. The purpose of these experiments is to demonstrate, as proof of concept, that the method is able to model relevant statistical properties in high dimensional data.

\subsection{Novelty detection}

Novelty detection, recognition of patterns in test data that are unlikely under the distribution of the training data, is a common unsupervised task for which density estimation is used. Here, we considered the problem of detecting mislabelled images of handwritten digits given a training set of correctly labelled images from the USPS dataset. The dataset consists of 7291 training and 2007 test images, each of which is a labelled, $16\times 16$ gray-scale image of a handwritten digit (0-9).

We modelled the joint distribution of the image (a 256 dimensional vector) and the label (a ten dimensional sparse vector, where the $l^{\mbox{th}}$ element is $1$ and the rest are $0$'s if the label is $l$) by a kernel \studentt density (\emph{\ourmethod}). We used a squared exponential kernel with length-scale chosen to be the median distance between data-points along digit dimensions and length-scale $1$ along the label dimensions. We trained the algorithm on a randomly selected subset of available training data and calculated predictive probabilities on a test dataset composed of 100 correctly labelled and 100 mislabelled test points. Digit-label pairs with predictive probability under a threshold were considered \emph{mislabelled}. We compared the performance of our method to two baseline methods, kernel density estimation (\emph{kde}, also called Parzen window estimate) and Dirichlet process mixtures of Gaussians (\emph{dpm}), based on the area under the receiver operating characteristic curve (AuC) metric. Hyper-parameters of all three methods were chosen by grid search so that the average AuC was maximised on a separate validation set. Table \ref{tab:digits_results} summarises the results of this comparison based on ten randomised iteration of the experiment with 2000 training and 200 test samples. We found that \emph{\ourmethod} outperformed both competing methods significantly ($p<0.01$, two sample T-test). The performance of \emph{dpm} was substantially worse than that of the other two, which demonstrates that general mixture models are very inefficient in modelling high dimensional complex densities. To investigate the difference between the performance of \emph{kde} and \emph{\ourmethod}, we carried out a second sequence of experiments where the size of the training set ranged from 10 to 400. Fig.~\ref{fig:AUC_and_Confusion} shows average AuC values as a function of training set size. We observed that the two algorithms perform similarly for small amounts of training data, the difference in performance becomes significant ($p<0.05$) for 60 training points or more.

\subsection{Label reconstruction\label{sec:classification}}

In these experiments, we considered learning the classification of handwritten digits. In order to assess the density modelling capabilities of our method, we posed this problem as an image reconstruction problem \footnote{We could have considered more general image reconstruction tasks, but reconstructing labels equips us with intuitive measures of loss, such as confusion and AuC.}. Again, we augmented the gray-scale image with 10 additional `pixels' representing a sparse coding of the labels 0-9. We first trained our density model on the augmented images on randomly selected training subsets of various sizes. Then, on a separate subset of test images we computed the probability of each label 0-9 conditioned on the image by computing the joint probability of the image with that label and renormalising. For each image we assigned the label for which the conditional predictive probability was the highest (maximum \emph{a posteriori} reconstruction). Results of comparison to \emph{kde} and \emph{DPM} with 2000 training and 400 test points are shown in \figref{fig:AUC_and_Confusion}. As measure of performance, we used average confusion, \ie the fraction of misclassified test images. We again found that \emph{kst} outperformed both other methods, and now the advantage of our method compared to \emph{kde} is more substantial. Fig.~\ref{fig:AUC_and_Confusion} shows confusion of \emph{\ourmethod} and \emph{kde} as a function of the number of training examples. The difference between the two methods becomes significant ($p<0.01$) for 60 training examples or more.


\subsection{Relative novelty detection}

\begin{figure*}
 \begin{center}
  \resizebox{\textwidth}{!}{
  \begin{tikzpicture}
    \node[anchor = south west, inner sep = 0in] at (0in,8pt) {\includegraphics[width=2.1in,keepaspectratio]{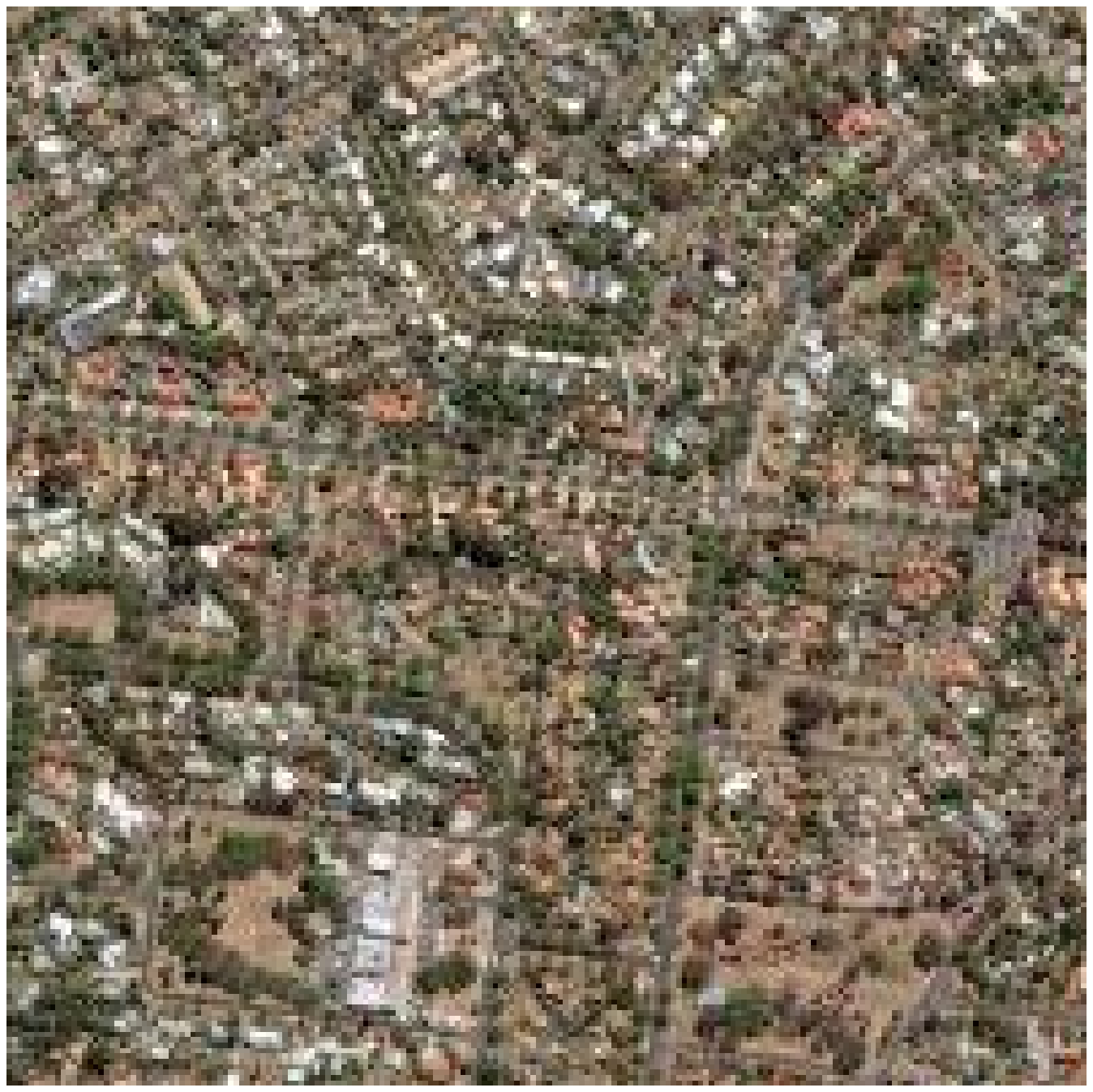}};
    \node[inner sep = 0in] at (1.05in,2.3in) {\textbf{A}: Background};
    \node[anchor = south west, inner sep = 0in] at (2.15in,8pt) {\includegraphics[width=2.1in,keepaspectratio]{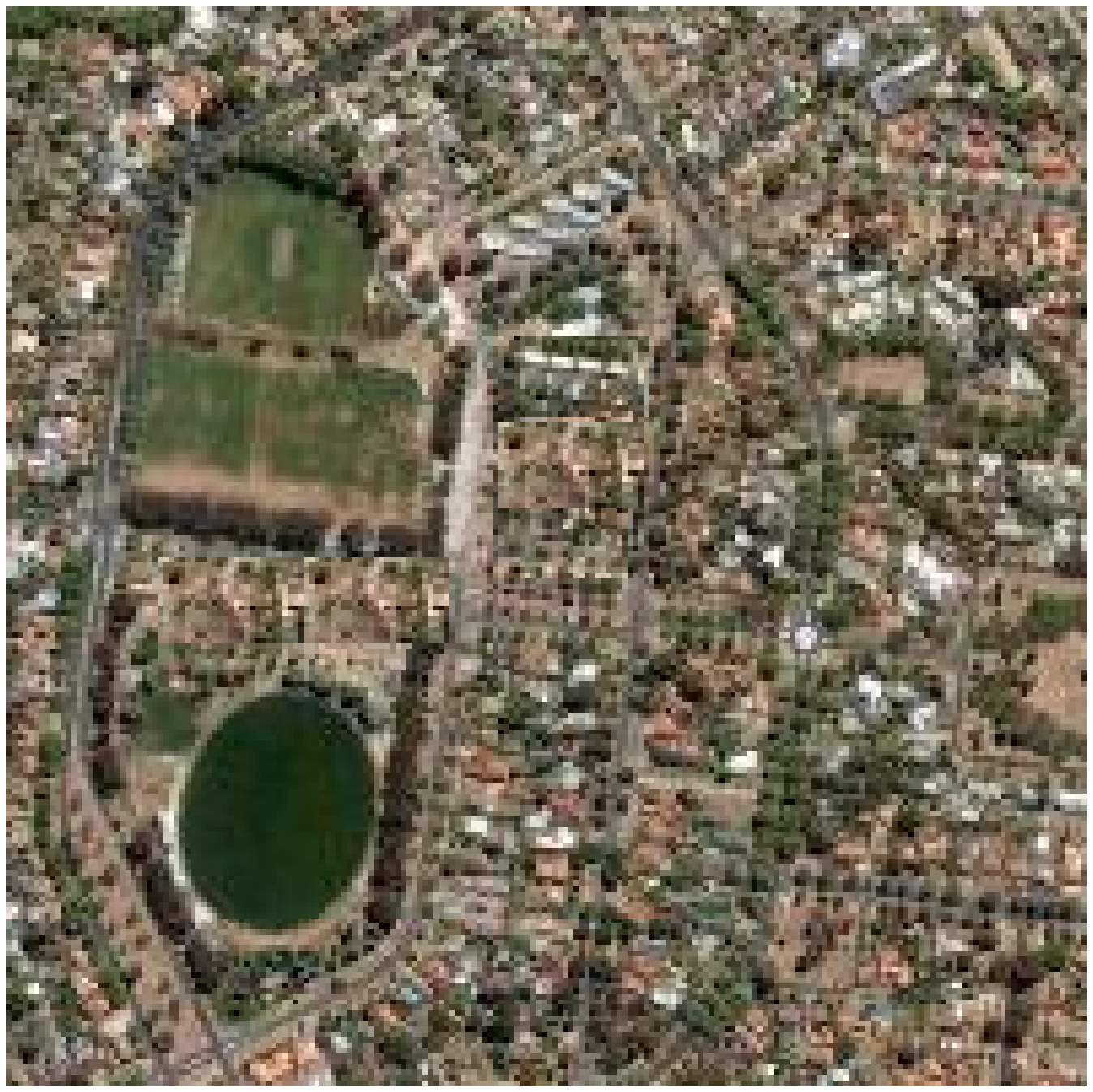}};
    \node[inner sep = 0in] at (3.2in,2.3in) {\textbf{B}: Target};
    \node[anchor = south west, inner sep = 0in] at (4.3in,8pt) {\includegraphics[width=2.1in,keepaspectratio]{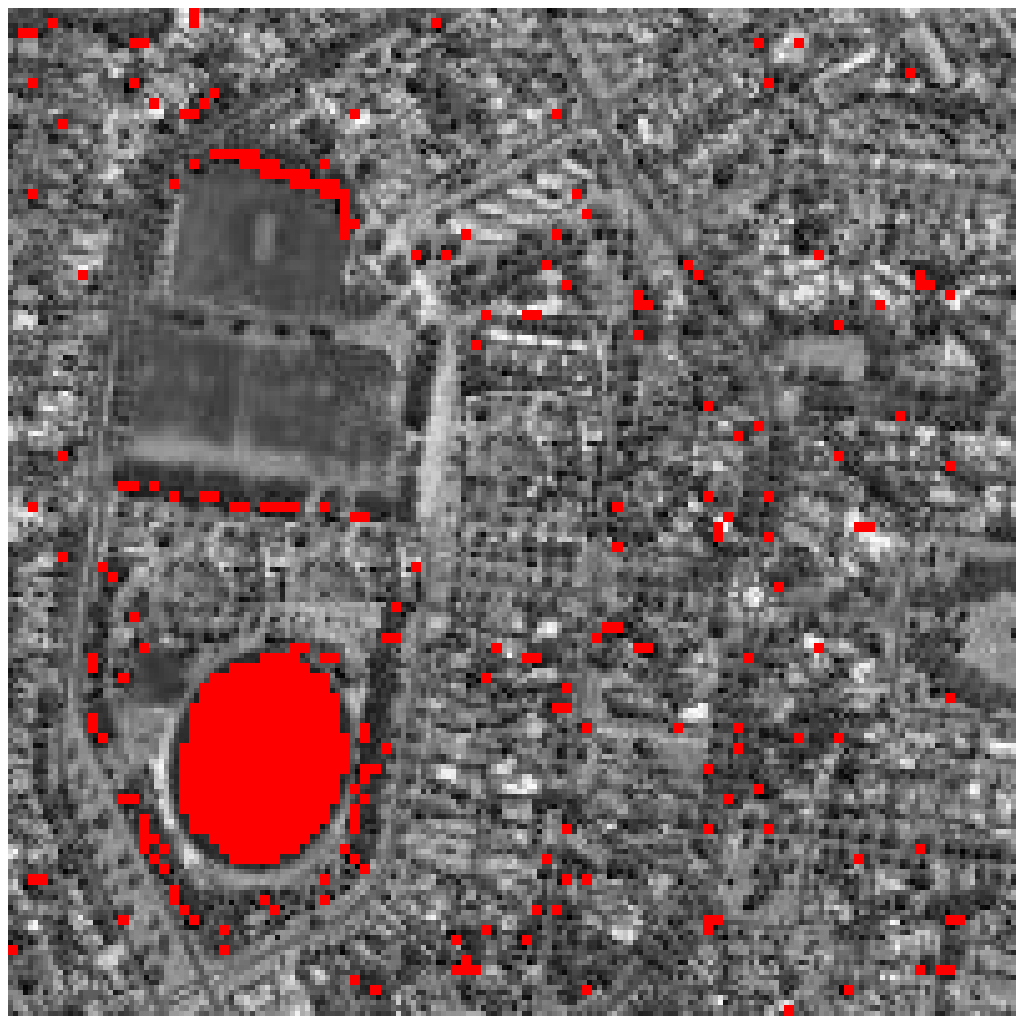}};
    \node[inner sep = 0in] at (5.35in,2.3in) {\textbf{C}: Relative novelty};
  \end{tikzpicture}
    }
 \end{center}

 \caption{\textbf{A,\ }Background and \textbf{B,\ }target images for relative novelty detection. \textbf{C,\ }Top 5\% novel patches in the target image are highlighted in \emph{red}. The results are qualitatively similar to those presented in~\cite[Fig. 2]{Smola2009}\label{fig:satellite}.}
\end{figure*}

Finally, we considered the problem of relative novelty detection~\cite{Smola2009}, for which we re-used the experimental setup in~\cite{Smola2009}. The data consists of two $200\times 200$ color satellite images, the \emph{background} and the \emph{target} (Fig.~\ref{fig:satellite}.A--B). The task is to detect novel objects on the target image relative to the background image. Treating RGB values in each non-overlapping $2\times 2$ patches as data points, we estimated the density over such patches in both the background ($\hat{q}(\bm{x})$) and in the target ($\hat{p}(\bm{x})$) images. A patch $\bm{x}$ was identified as relatively novel if the ratio $\hat{p}(\bm{x})/\hat{q}(\bm{x})$ was above a certain threshold. The results produced by our algorithm are similar to that obtained by using state of the art novelty detection algorithms (c.\,f.\ Fig.~2 in~\cite{Smola2009}), even though here we only used a 3\% random subsample of available image patches to estimate target and background densities.

\section{Discussion}

\paragraph{Summary} In this paper, we considered applying the kernel trick as a tool for constructing nonparametric Bayesian methods, and suggested a general recipe for obtaining novel analytically tractable \emph{Bayesian kernel machines}. The starting point is a suitably simple parametric Bayesian model in which inference is tractable. We have demonstrated how arguments based on orthonormal invariance can be used to guide our choice of models that are amenable to kernel trick. Having defined the basic model, one needs to express posterior predictive distributions in terms of kernel evaluations, then apply the kernel trick in those equations. By using kernels with infinite dimensional feature spaces, novel tractable nonparametric Bayesian methods may be obtained in this way.

To illustrate our general approach, we studied the problem of density estimation and presented kernel \studentt density estimation, a novel Bayesian kernel machine for density modelling. We followed our general methodology to arrive at a closed form expression that is exact up to a multiplicative constant. We also carried out numerical experiments to investigate the performance of the method. The results demonstrate that the method is capable of estimating the distribution of high dimensional real-world data, and can be used in various applications such as novelty detection and reconstruction. The most apparent limitation of the kernel \studentt density model is the intractability of its normalisation constant. Nevertheless, the unnormalised density can still be used in a variety of unsupervised learning tasks, such as image reconstruction or novelty detection. As other kernel methods, our algorithm has cubic complexity in the number of training points, though sparse approximations~\cite{Fine2001} can be used to construct faster algorithms for large-scale applications. Another advantage of the kernel approach is that the expressions can be formally extended to define non-trivial discrete distributions over \eg graphs, permutations or text.

\paragraph{Conclusions and future work} We believe that the present paper opens the door for developing a novel class of Bayesian kernel machines. The methodology presented in this paper can be applied in further problems to obtain nonparametric methods based on parametric Bayesian models. Investigating merits and limitations of this general approach in various statistical estimation and decision problems is certainly an interesting theme for future research. One particularly exciting direction that we plan to pursue is a Bayesian two-sample hypothesis test analogous to the frequentist test based on characteristic kernels~\cite{Sriperumbudur2008}. For hypothesis testing, one needs to apply the kernel trick to a \emph{ratio} of marginal likelihoods, rather than to a predictive density, which enables cancellations which simplify the issues with normalisation or Jacobians. Other potentially fruitful directions include semi-supervised regression and Bayesian numerical analysis~\cite{OHagan1992}.

\paragraph{Acknowledgements} We would like to thank Zoubin Ghahramani, Carl Rasmussen, Arthur Gretton, Peter Orbanz
and Le Song for helpful discussions. In particular, the proof in
Appendix \ref{sec:Jacobian_shift} is mainly due to Le Song. Part of this work was supported
under the EPSRC grant EP/F026641/1. Ferenc Husz\'{a}r is supported by Trinity College Cambridge.

\bibliographystyle{nature3}
\bibliography{bibliography/Bayesian_kernel}

\appendix

\twocolumn[
\section{Change-of-variable formula\label{sec:change_of_variable}}

\paragraph{Theorem} Let $\mathcal{X}$ and $\mathcal{F}$ be $d$ and $D$ dimensional Euclidean spaces respectively ($D>d$), ${\varphi:\mathcal{X} \mapsto \mathcal{F}}$ be a smooth injective mapping and let $q(\bm{\phi})$ define a density function with respect to the Lebesgue measure on $\mathcal{F}$. Then the restriction of distribution $q$ to the observation manifold $\varphi(\mathcal{X})$ (\ie conditioning on the event of being in the observation manifold) induces the following density in the input space $\mathcal{X}$:
\begin{equation} \label{eq:densityChange}
p(\bm{x}) \propto q(\bm{\varphi}(\bm{x})) \cdot \det\left( \left\langle\frac{\partial\varphi(\bm{x})}{\partial x^i} ,  \frac{\partial\varphi(\bm{x})}{\partial x^j}\right\rangle \right)^{\frac{1}{2}}_{i,j=1,\ldots, d}\mbox{.}
\end{equation}

We suspect this result may have appeared in several forms in the literature, but we didn't succeed to find a citation, and so we provide a proof here.

\paragraph{Proof} Note that we cannot apply the standard change-of-variable formula directly as the two spaces have different dimensions. So we augment the space $\mathcal{X}$ with an Euclidean space $\Xi$ of dimension $D-d$. We then define the following mapping $F:\mathcal{X}\times\Xi \mapsto \mathcal{F}$:
$$
F(\bm{x},\bm{\xi}) = \varphi(\bm{x}) + V_{\bm{x}} \bm{\xi},
$$
where $V_{\bm{x}}$ is a $D\times(D-d)$ matrix of mutually orthonormal vectors to the tangent space of the manifold at $\bm{x}$ (\ie a basis for the orthogonal space to the tangent space at $\bm{x}$). For small enough $\bm{\xi}$'s (which could depend on $\bm{x}$), $F$ will map small neighborhoods of $(\bm{x},\bm{0})$ to neighborhoods of $\varphi(\bm{x})$ in an injective manner. We thus apply the standard change of variable formula to map the density $q(\bm{\phi})$ to a density on an open set around $\mathcal{X}\times\{\bm{0}\}$:
$$
    p(\bm{x},\bm{\xi}) = q(F(\bm{x},\bm{\xi})) \left\vert\det J_F(\bm{x},\bm{\xi}) \right\vert ,
$$
where $J_F(\bm{x},\bm{\xi})$ is the Jacobian matrix of the $F$ mapping. To condition on being in the observation manifold, we simply need to condition on $\bm{\xi}=0$, and so the density we want to compute is:
$$
    p(\bm{x}) \propto p(\bm{x},\bm{0}) = q(F(\bm{x},\bm{0})) \left\vert\det J_F(\bm{x},\bm{0}) \right\vert .
$$
Now, $J_F(\bm{x},\bm{0})=(J_\varphi(\bm{x}), V_{\bm{x}})$ where $J_\varphi(\bm{x})=(\frac{\partial\varphi(\bm{x})}{\partial x^1}, \ldots, \frac{\partial\varphi(\bm{x})}{\partial x^d})$ is the Jacobian matrix of $\varphi$. Finally, we use the fact that $\vert\det(J_F)\vert$ = $(\det(\transpose{J_F} J_F))^{1/2}$ for any squared matrix $J_F$ to get that:
$$
\left\vert\det\left(J_F (\bm{x},\bm{0})\right)\right\vert = \det\left(\begin{array}{cc}
                        \transpose{J_\varphi(\bm{x})}J_\varphi(\bm{x}) & \bm{0} \\
                        \bm{0} & \bm{I}
                      \end{array} \right)^{\frac{1}{2}} = \det\left( \transpose{J_\varphi(\bm{x})}J_\varphi(\bm{x}) \right)^{\frac{1}{2}},
$$
where we used that $\transpose{J_\varphi(\bm{x})} V_{\bm{x}}=0$ and $\transpose{V_{\bm{x}}} V_{\bm{x}}=\bm{I}$ by orthonormality of the $V_{\bm{x}}$ with the tangent space spanned by the columns of $J_\varphi(\bm{x})$. Finally, we use the fact that:
$$
 \transpose{J_\varphi(\bm{x})}J_\varphi(\bm{x}) = \left( \left\langle\frac{\partial\varphi(\bm{x})}{\partial x^i} ,  \frac{\partial\varphi(\bm{x})}{\partial x^j}\right\rangle \right)_{i,j=1,\ldots, d}
$$
to obtain~\eqref{eq:densityChange}. $\square$
]

\twocolumn[
\section{Jacobian term for translation invariant kernel\label{sec:Jacobian_shift}}

We provide the sketch of a proof that $\left\langle\frac{\partial\varphi(\bm{x})}{\partial x^i} ,  \frac{\partial\varphi(\bm{x})}{\partial x^j}\right\rangle$ doesn't depend on $\bm{x}$ for a translation invariant kernel under suitable regularity conditions.

\paragraph{}By Bochner's theorem (\eg theorem 4.1 in~\cite{Rasmussen2006}), a translation invariant kernel can be expressed as the inverse Fourier transform of a positive finite measure. If moreover we assume that the Fourier transform of the kernel exists, this measure will have a density $p(\bm{s})$ with respect to the Lebesgue measure, and so we can write:
\begin{align*}
 k(\bm{x}-\bm{y}) & =  \int_{\mathbb{R}^d} p(\bm{s}) e^{2\pi i \transpose{\bm{s}} (\bm{x}-\bm{y})} d\bm{s} \\
 &= \int p(\bm{s}) \left( \cos(2\pi \transpose{\bm{s}} (\bm{x}-\bm{y}))+i
 \sin(2\pi \transpose{\bm{s}} (\bm{x}-\bm{y})) \right) d\bm{s} \\
 &= \int p(\bm{s}) \left( \cos(2\pi \transpose{\bm{s}} (\bm{x}-\bm{y})) \right) d\bm{s}
\end{align*}
as $\sin(\transpose{\bm{s}}\bm{\delta})$ is an odd function of $\bm{s}$ for any $\bm{\delta}$ and $p(\bm{s})$ is a symmetric function given that it is the inverse Fourier transform of the symmetric function $k$, so the Cauchy principal value of the second term vanishes. Now using a trigonometric identity, we get:
\begin{align*}
 k(\bm{x}-\bm{y}) & = \int p(\bm{s}) \left( \left( \cos(2\pi \transpose{\bm{s}}\bm{x})  \cos(2\pi \transpose{\bm{s}}\bm{y}) \right) +
 \left( \sin(2\pi \transpose{\bm{s}}\bm{x})  \sin(2\pi \transpose{\bm{s}}\bm{y}) \right) \right) d\bm{s} \\
  &= \int \left\langle \sqrt{p(\bm{s})} \left( {\cos(2\pi \transpose{\bm{s}}\bm{x}) \atop \sin(2\pi \transpose{\bm{s}}\bm{x})} \right) , \sqrt{p(\bm{s})} \left( {\sin(2\pi \transpose{\bm{s}}\bm{y}) \atop \sin(2\pi \transpose{\bm{s}}\bm{y})} \right) \right\rangle d\bm{s}
\end{align*}
So we can obtain an explicit feature mapping by defining:
$$
\varphi(\bm{x}) = \left( \sqrt{p(\bm{s})} \left( {\cos(2\pi \transpose{\bm{s}}\bm{x}) \atop \sin(2\pi \transpose{\bm{s}}\bm{x})} \right) \right)_{\bm{s} \in \mathbb{R}^d}.
$$
Assuming that $p(\bm{s})$ goes to zero sufficiently quickly, we have that:
$$
\frac{\partial \varphi(\bm{x})}{\partial x_k} = \left( 2\pi s_k \sqrt{p(\bm{s})} \left( {-\sin(2\pi \transpose{\bm{s}}\bm{x}) \atop \cos(2\pi \transpose{\bm{s}}\bm{x})} \right) \right)_{\bm{s} \in \mathbb{R}^d}
$$
and using a similar derivation as we used above, we get that:
\begin{align*}
\left\langle\frac{\partial\varphi(\bm{x})}{\partial x_k} ,  \frac{\partial\varphi(\bm{x})}{\partial x_l}\right\rangle &= 4\pi^2 \int
s_k s_l p(\bm{s}) e^{2\pi i \transpose{\bm{s}} (\bm{x}-\bm{x})} d\bm{s} \\
&=  4\pi^2 \int
s_k s_l p(\bm{s}) e^{0} d\bm{s}
\end{align*}
which doesn't depend on $\bm{x}$ as we wanted to show. $\square$
]
\end{document}